\documentclass[pdflatex,sn-mathphys-num]{sn-jnl}

\usepackage{graphicx}%
\usepackage{multirow}%
\usepackage{amsmath,amssymb,amsfonts}%
\usepackage{amsthm}%
\usepackage{mathrsfs}%
\usepackage[title]{appendix}%
\usepackage{xcolor}%
\usepackage{textcomp}%
\usepackage{manyfoot}%
\usepackage{booktabs}%
\usepackage{algorithm}%
\usepackage{algorithmicx}%
\usepackage{algpseudocode}%
\usepackage{listings}%
\usepackage{rotating} 
\usepackage{array}    
\newcolumntype{L}[1]{>{\raggedright\arraybackslash}p{#1}}
\usepackage{adjustbox}
\usepackage{longtable}

\theoremstyle{thmstyleone}%
\theoremstyle{thmstyletwo}%

\theoremstyle{thmstylethree}%

\usepackage[justification=centering]{caption}

\usepackage{booktabs,tabularx,array}
\newcolumntype{L}[1]{>{\raggedright\arraybackslash}p{#1}}
\renewcommand{\arraystretch}{0.85}
\begin{document}

\title[Article Title]{Towards Transparent AI: A Survey on Explainable Language Models}


\author[1]{\fnm{Avash} \sur{Palikhe}}\email{apali007@fiu.edu}

\author[1]{\fnm{Zichong} \sur{Wang}}\email{ziwang@fiu.edu}

\author[1]{\fnm{Zhipeng} \sur{Yin}}\email{zyin007@fiu.edu}

\author[2]{\fnm{Rui} \sur{Guo}}  

\author[3]{\fnm{Qiang} \sur{Duan}}

\author[4]{\fnm{Jie} \sur{Yang}}

\author*[1]{\fnm{Wenbin} \sur{Zhang}}\email{wenbin.zhang@fiu.edu}

\affil[1]{\orgname{Florida International University}, \city{Miami}, \country{United States}}
\affil[2]{\orgname{University of Florida}, \orgaddress{\street{Gainesville}, \country{United States}}}
\affil[3]{\orgname{Pennsylvania State University}, \orgaddress{\street{Montgomery County}, \country{United States}}}
\affil[4]{\orgname{University of Wollongong}, \orgaddress{\street{Wollongong}, \country{Australia}}}


\abstract{
Language Models (LMs) have significantly advanced natural language processing and enabled remarkable progress across diverse domains, yet their black-box nature raises critical concerns about the interpretability of their internal mechanisms and decision-making processes. This lack of transparency is particularly problematic for adoption in high-stakes domains, where stakeholders need to understand the rationale behind model outputs to ensure accountability. On the other hand, while explainable artificial intelligence (XAI) methods have been well studied for non-LMs, they face many limitations when applied to LMs due to their complex architectures, considerable training corpora, and broad generalization abilities. Although various surveys have examined XAI in the context of LMs, they often fail to capture the distinct challenges arising from the architectural diversity and evolving capabilities of these models. To bridge this gap, this survey presents a comprehensive review of XAI techniques with a particular emphasis on LMs, organizing them according to their underlying transformer architectures: encoder-only, decoder-only, and encoder-decoder, and analyzing how methods are adapted to each while assessing their respective strengths and limitations. Furthermore, we evaluate these techniques through the dual lenses of plausibility and faithfulness, offering a structured perspective on their effectiveness. Finally, we identify open research challenges and outline promising future directions, aiming to guide ongoing efforts toward the development of robust, transparent, and interpretable XAI methods for LMs.
}




\maketitle

\section{Introduction}

In the evolving field of natural language processing (NLP), language models (LMs) such as BERT~\cite{devlin2018bert}, T5~\cite{raffel2020exploring}, GPT-4~\cite{openai2023gpt4}, and LLaMA-2~\cite{touvron2023llama} have demonstrated remarkable progress across a wide range of tasks and application domains, including machine translation, code generation, medical diagnosis, and personalized education~\cite{lee2020biobert,chen2021evaluating,ng2024educational,lopes2020litetrainingstrategiesportugueseenglish}. Despite these advancements, LMs inherently exhibit a black-box nature due to their considerable parameter scales, extensive training data, and complex architectures, making it challenging to understand their internal mechanisms and decision-making processes~\cite{schwartz2024black}. This lack of transparency introduces several critical limitations, including diminished trust, reduced accountability, and increased difficulty in identifying biases or debugging errors in model outputs. Consequently, these challenges become particularly critical when LMs are deployed in high-stakes domains such as healthcare, finance, and law, where transparency and accountability are essential because errors can lead to serious consequences~\cite{10.1145/3604237.3626869,schwartz2024black,shu2024lawllm}.

On the other hand, explainable artificial intelligence (XAI) methods developed for transparency and accountability in non-LM domains have been extensively studied~\cite{jung2021explaining, papanastasopoulos2020explainable, freeborough2022investigating, burkart2021survey}. For example, feature attribution techniques such as Integrated Gradients~\cite{sundararajanIntegratedGradients} are widely adopted in non-LM settings to interpret model predictions by quantifying the contribution of individual input features. However, applying such methods directly to LM is difficult due to their considerable parameter scale and broad generalization abilities~\cite{kokalj2021bert}. This has motivated researchers to design XAI approaches tailored to LMs. For instance, Chain-of-Thought (CoT) prompting~\cite{wei2022chain, kojima2022large} exposes intermediate reasoning steps to improve transparency, while self-explanation methods generate human-readable justifications alongside predictions~\cite{juraska2021attention, cao2021attention}. Although various existing surveys have examined XAI in the context of LMs, they often overlook the architecture-specific challenges that arise from the diverse combinations of transformer encoder and decoder components~\cite{zhao2023explainability, luo2024survey, zhang2022surveyxai, 10.1145/3561048,islam2021explainableartificialintelligenceapproaches}. Since each architecture processes and represents information differently~\cite{wang2025history}, these variations lead to distinct explainability issues. Consequently, a unified framework is essential to systematically examine XAI methods across model architectures and to address the unique interpretability challenges posed by LMs.

To this end, this survey provides a systematic review of XAI methods for language models, organizing them by their underlying transformer architectures, discussing their evaluation mechanisms, and identifying open challenges along with future research directions. \textit{To the best of our knowledge, this is the first survey to present a comprehensive framework of XAI methods specifically tailored to LMs, introducing a standardized taxonomy based on transformer architectures, analyzing architecture-specific challenges, and assessing existing evaluation strategies}. By comparing explainability methods from an architecture-aware perspective, we offer a holistic view of current approaches and their effectiveness. The main contributions of this paper are as follows:

\begin{itemize}
    \item Proposes a taxonomy of XAI methods for LMs structured around their underlying transformer architectures.  
    \item Provides a comparative analysis of these methods, highlighting key features and limitations across different design choices.  
    \item Explore evaluation strategies for explanations, emphasizing reliability through the dual perspectives of faithfulness and plausibility.  
\end{itemize}

\section{Background}

\subsection{Language Models}

LMs have become fundamental to modern NLP, driving significant progress in applications such as machine translation, dialogue generation, question answering, and code synthesis~\cite{brown2020language, openai2023gpt4}. Most LMs are built upon transformer architectures~\cite{vaswani2017attention}, which leverage self-attention mechanisms to capture contextual dependencies. Based on their architectural design, LMs can be broadly categorized into three groups: encoder-only, decoder-only, and encoder-decoder models, each with distinct features and applications as detailed below.

Encoder-only models, such as BERT~\cite{devlin2018bert} and RoBERTa~\cite{liu2019robertarobustlyoptimizedbert}, consist solely of the encoder component and process input using bidirectional self-attention. They are pretrained with the masked language modeling objective and excel at tasks requiring text comprehension~\cite{minaee2025largelanguagemodelssurvey}. In contrast, decoder-only models, including GPT-4~\cite{openai2023gpt4} and LLaMA-2~\cite{touvron2023llama}, consist exclusively of the decoder component and generate text autoregressively via causal self-attention~\cite{KALYAN2022103982}. By predicting each token conditioned on the preceding tokens, these models excel at tasks involving free-form generation, such as dialogue, storytelling, and code synthesis~\cite{wang2025history}. Finally, encoder-decoder models, such as T5~\cite{raffel2020exploring} and BART~\cite{lewis2019bartdenoisingsequencetosequencepretraining}, integrate both components connected through cross-attention. Trained on sequence-to-sequence objectives, they are well-suited for tasks requiring both understanding and generation, including machine translation and summarization~\cite{minaee2025largelanguagemodelssurvey}.

\subsection{Explainable Artificial Intelligence (XAI)}

Explainable Artificial Intelligence (XAI) seeks to make AI systems more interpretable and transparent by providing human-understandable insights into their internal reasoning and decision-making processes~\cite{luo2024survey}. As AI continues to be deployed in high-stakes domains such as healthcare, finance, education, and law, the ability to explain predictions has become critical to building trust, ensuring accountability, and meeting regulatory requirements~\cite{lee2020biobert,chen2021evaluating,ng2024educational,lopes2020litetrainingstrategiesportugueseenglish}. This growing demand for interpretability has fueled the rapid development of XAI methods and frameworks.

Within this context, non-LMs such as random forests or neural networks typically possess relatively simpler architectures and have been more extensively studied from an explainability perspective~\cite{gulowaty2021extracting}. These models generally process shorter contexts and utilize simpler feature representations, which makes their decision-making mechanisms easier to analyze. Consequently, a wide range of XAI techniques have been proposed and successfully applied to these models~\cite{jung2021explaining, papanastasopoulos2020explainable, freeborough2022investigating, burkart2021survey}. For instance, methods, such as Integrated Gradients~\cite{sundararajanIntegratedGradients} approximate model predictions by estimating feature contributions, while counterfactual explanation methods identify minimal input perturbations required to alter a model’s decision, offering interpretable insights into the factors influencing predictions~\cite{dai2022counterfactual}. Due to the simple structure and limited contextual complexity of these models, such approaches tend to yield explanations that are generally more faithful and human-interpretable compared to LMs.

\subsection{XAI for LMs}

Unlike traditional non-LM models with relatively simpler architectures, language models pose unique explainability challenges due to their transformer-based designs and massive parameter scales~\cite{minaee2025largelanguagemodelssurvey}. Existing XAI techniques developed for classical models often fail to transfer directly to LMs, as they cannot adequately capture attention flows, contextual embeddings, or emergent reasoning behaviors~\cite{kokalj2021bert}. Moreover, LMs are built on diverse transformer architectures, encoder-only, decoder-only, and encoder–decoder, each defined by distinct attention mechanisms and pre-training objectives~\cite{wang2025history}. These architectural differences necessitate tailored approaches, making the design of effective XAI techniques for LMs complex and essential.

Moreover, the tension between plausibility and faithfulness is a well-known challenge in explainable artificial intelligence across many types of models~\cite{zhao2023explainability}. For classical models such as random forests or neural networks, XAI techniques often generate explanations that appear plausible to humans but do not always faithfully reflect the model’s actual decision process~\cite{pmlr-v235-chen24bl}. In language models, this problem is amplified by their transformer based architectures, massive parameter scales, and emergent reasoning behaviors~\cite{minaee2025largelanguagemodelssurvey}. Explanations that seem convincing may overlook the true underlying dynamics of attention flows, contextual embeddings, or reasoning chains. This distinction is especially critical in high stakes applications such as healthcare, finance, and law, where misleading yet plausible explanations can compromise trust, accountability, and safety~\cite{10.1145/3583558}. Addressing this challenge requires XAI methods explicitly tailored to LMs architectures, alongside rigorous evaluation frameworks that jointly assess plausibility and faithfulness, ensuring explanations are both human interpretable and faithful to the model’s internal reasoning~\cite{deyoung2020eraser}.

\section{Taxonomy}

\begin{figure}[!htb]
    \centering
     \includegraphics[width=0.83\textwidth]{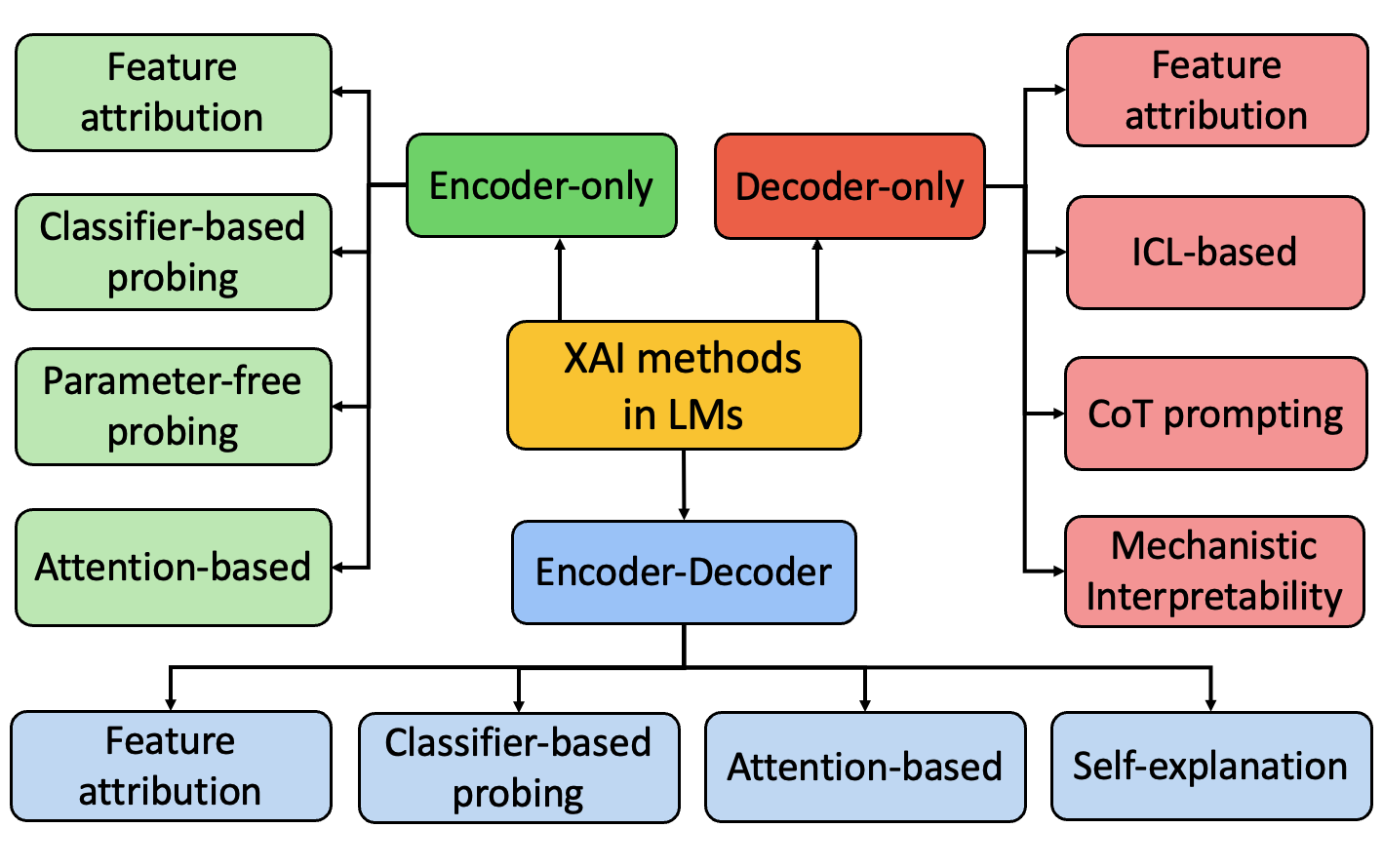}
    \vspace{+0.4cm}
    \caption{The proposed taxonomy for XAI in language models.}
    \label{fig:taxonomy}
\end{figure}

To address the diverse explainability challenges posed by language models and to provide actionable guidelines, we present a novel taxonomy for categorizing XAI methods based on their underlying transformer architectures. This taxonomy not only structures existing work but also offers practical guidance for selecting and developing methods suited to different model families. As illustrated in Figure~\ref{fig:taxonomy}, the classification organizes approaches into three groups. First, XAI methods for encoder-only models interpret contextual embeddings by leveraging bidirectional self-attention to analyze internal representations. Second, XAI methods for decoder-only models frequently rely on prompt engineering and in-context learning strategies to reveal the reasoning processes of these autoregressive models with massive parameter scales. Third, XAI methods for encoder-decoder models provide interpretability at both encoding and decoding stages, exploiting cross-attention to trace the flow of information across components. By standardizing XAI methods along these architectural lines, the taxonomy enables meaningful comparisons of their mechanisms and characteristics, highlights architecture specific challenges, and clarifies how explanatory effectiveness can be evaluated.

\section{Overview}

Based on the proposed taxonomy, this section presents an overview of XAI methodologies for language models, as summarized in Table~\ref{tab:xai-methods}. The table organizes prior studies across six key dimensions: (1) \textbf{Method}, indicating the specific XAI technique used (\textit{e.g.}, LIME, TransSHAP, ROME); (2) \textbf{Architecture (Arch.)}, specifying the applicable model type, encoder-only, decoder-only, or encoder–decoder; (3) \textbf{Type}, denoting the broader class of approach, such as feature attribution, ICL-based, attention-based, or self-explanation; (4) \textbf{Model}, listing the representative language model (\textit{e.g.}, BERT, GPT-2, T5); (5) \textbf{Scope}, distinguishing whether the explanation is local (L) to individual predictions or global (G) to overall model behavior; and (6) \textbf{Integration (Inte.)}, identifying when interpretability is introduced, either post hoc (PH) or at prompt time (PT). This structured comparison facilitates a deeper understanding of methodological distinctions, architectural compatibility, and interpretability timing, thereby supporting informed evaluation and selection of XAI techniques for various LM settings.

\renewcommand{\arraystretch}{1.2} 

\begin{longtable}{|p{1cm}| p{2.7cm} |p{1.7cm}| p{2.3cm}| p{1.8cm} |p{1.0cm}| p{1.0cm}|}
\caption{Summary of different methodologies used for explainable AI (XAI) in language models (LMs). 
Abbreviations: Arch. = Architecture, Inte. = Integration, L = Local, G = Global, PH = Post-hoc, PT = Prompt-time.}
\label{tab:xai-methods} \\

\hline
\textbf{Paper} & \textbf{Method} & \textbf{Arch.} & \textbf{Type} &
\textbf{Model} & \textbf{Scope} & \textbf{Inte.} \\
\hline
\hline
\endfirsthead

\hline
\textbf{Paper} & \textbf{Method} & \textbf{Arch.} & \textbf{Type} &
\textbf{Model} & \textbf{Scope} & \textbf{Inte.} \\
\hline
\endhead

\hline
\multicolumn{7}{r}{\small Continued on next page} \\
\endfoot

\hline

\endlastfoot

\cite{szczepanski2021new} & LIME & Encoder-only & Feature attribution-based & BERT & L & PH \\
\hline
\cite{lundberg2017shap} & TransSHAP & Encoder-only & Feature attribution-based & BERT & L & PH \\
\hline
\cite{tenney2019you} & Edge probing & Encoder-only & Classifier-based probing & BERT & L, G & PH \\
\hline
\cite{tenney2019you} & Token representation probing & Encoder-only & Classifier-based probing & BERT & L & PH \\
\hline
\cite{wu2020perturbed} & Perturbation masking & Encoder-only & Parameter-free probing & BERT & G & PH \\
\hline
\cite{wu2020perturbed} & DIRECTPROBE & Encoder-only & Parameter-free probing & BERT & G & PH \\
\hline
\cite{vig2019analyzing} & BertViz & Encoder-only & Attention-based & BERT & L & PH \\
\hline
\cite{hoover-etal-2020-exbert} & ExBERT & Encoder-only & Attention-based & BERT & L & PH \\
\hline
\cite{barkan2024llm} & AML & Decoder-only & Feature attribution-based & LLaMA-2 7B, Mistral 7B & L & PH \\
\hline
\cite{kariyappa2024progressive} & Progressive Inference & Decoder-only & Feature attribution-based & GPT-2, LLaMA-2 & L & PH \\
\hline
\cite{wei2023larger} & ICL with priors \& mappings & Decoder-only & ICL-based analysis & GPT-3, PaLM & G & PH \\
\hline
\cite{li2023saliencyicl} & Saliency-Guided ICL Analysis & Decoder-only & ICL-based analysis & Instruct-GPT, GPT-2 & L & PH \\
\hline
\cite{zhang2024sea} & SEA-CoT & Decoder-only & CoT prompting & LLaMA-2 & L & PT \\
\hline
\cite{zhou2022least} & Least-to-Most Prompting & Decoder-only & CoT prompting & GPT-3 & L & PT \\
\hline
\hline
\cite{meng2022locating} & ROME & Decoder-only & Mechanistic interpretability & GPT-2 XL & L & PH \\
\hline
\cite{conmy2023towards} & ACDC & Decoder-only & Mechanistic interpretability & GPT-2 Small & G & PH \\
\hline
\cite{liu2024reliabilityexplainabilitylanguagemodels} & Program generation explanations & Encoder-decoder & Feature attribution-based & T5, CodeT5, CodeReviewer, CodeT5+ & L & PH \\
\hline
\cite{enouen-etal-2024-textgenshap} & TextGenSHAP & Encoder-decoder & Feature attribution-based & T5-large, T5-XXL, T5-FiD & L & PH \\
\hline
\cite{koto-etal-2021-discourse} & Document-level discourse probing & Encoder-decoder & Classifier-based probing & T5, BART & G & PH \\
\hline
\cite{li-etal-2021-implicit} & Implicit meaning representation probing & Encoder-decoder & Classifier-based probing & T5, BART & G & PH \\
\hline
\cite{juraska2021attention} & Attention-Guided Decoding & Encoder-decoder & Attention-based & T5, BART & L & PH \\
\hline
\cite{cao2021attention} & Attention Head Masking & Encoder-decoder & Attention-based & BART, PEGASUS & G & PH \\
\hline
\cite{narang2020wt5} & WT5 & Encoder-decoder & Self-explanation & WT5-Base, WT5-11B & L & PT \\
\hline
\cite{yordanov2022few} & Few-Shot NLE Transfer & Encoder-decoder & Self-explanation & T5 & L & PT \\
\hline
\end{longtable}

\section{XAI Methods for LMs}
\label{sec:XAI_methods}
This section provides a comprehensive overview of XAI methods applied to LMs, categorized according to their underlying transformer architectures: encoder-only, decoder-only, and encoder-decoder. For encoder-only models, XAI approaches, such as feature attribution-based methods and classifier-based probing primarily focus on interpreting contextual embeddings and analyzing bidirectional self-attention mechanisms to explain model predictions. In contrast, decoder-only models adopt techniques such as chain-of-thought (CoT) prompting and in-context learning (ICL), which are designed to interpret the autoregressive generation process and often leverage prompt engineering strategies. Finally, encoder-decoder models utilize methods such as attention-based techniques and self-explanation approaches that aim to provide interpretability across both encoding and decoding stages by utilizing cross-attention mechanisms. The following sections discuss in detail the XAI approaches for each category.

\subsection{XAI Methods for Encoder-only LMs}

XAI methods for encoder-only models seek to provide insights into the internal mechanisms and decision-making processes of models that rely solely on an encoder architecture, such as BERT~\cite{devlin2018bert} and RoBERTa~\cite{liu2019robertarobustlyoptimizedbert}. These models employ a bidirectional self-attention mechanism to generate contextualized embeddings, making explanations particularly focused on how contextual information influences token representations~\cite{szczepanski2021new}. We discuss several key explainability approaches for these models, including feature attribution-based methods~\cite{szczepanski2021new}, classifier-based probing~\cite{tenney2019you}, parameter-free probing~\cite{wu2020perturbed} and attention-based methods~\cite{vig2019analyzing}.
\sloppy
\subsubsection{Feature Attribution-based Methods}
\label{feature-attribution-encoder-only}

Feature attribution-based methods evaluate the relevance of individual features, such as words or phrases, to a model’s prediction~\cite{zhao2023explainability}. They assign a score to each feature, interpreting its contribution to the output. These methods are particularly well-suited for encoder-only models, which are pre-trained with a masked language modeling (MLM) objective~\cite{devlin2018bert}. This objective encourages bidirectional context modeling, yielding stable and context-rich embeddings that can be reliably analyzed to assess the influence of each feature on the final prediction~\cite{wu2021explainingexplanationsbertempirical}.

One popular line of research extends the commonly used feature attribution technique, LIME, from non-LMs to LMs. Specifically, LIME explains a model’s predictions by approximating them locally with an interpretable model. Building on this technique, Szczepański et al.~\cite{szczepanski2021new} proposed an approach that adapts LIME to provide explainability for encoder-only LMs such as BERT, fine-tuned for fake news detection. In their method, it explains the predictions of the encoder-only model by highlighting words and their corresponding weights, which represent their impact on the prediction probability. The increase in weight for these words raises the likelihood of the sentence being classified as fake and vice versa. However, LIME has limitations because it may not satisfy additive attributions defined by Lundberg and Lee~\cite{lundberg2017shap}, such as local accuracy, consistency, and missingness. For example, it often fails to ensure local accuracy, as the sum of the feature contributions does not necessarily equal the model’s actual output for a given instance. Similarly, it doesn't guarantee consistency, where increasing a feature's importance in the model may not always increase its attribution score. Finally, LIME may not ensure missingness, as features that are absent or masked can still be assigned non-zero attribution scores, potentially leading to misleading interpretations. These shortcomings can reduce the reliability of LIME-based explanations, especially when applied to language models.

Some methods address these limitations by extending SHAP~\cite{lundberg2017shap} to language models. Specifically, SHAP is an XAI technique that introduces Shapley values as a unified measure of additive feature importance to explain the model decisions. To adapt this technique to encoder-only LMs, such as BERT, Kokalj et al.~\cite{kokalj2021bert} proposed a method called TransSHAP. They demonstrate the method in a tweet sentiment classification task, where TransSHAP explains BERT's predictions for positive and negative sentiment using visualization. The visualizer allows the user to compare the direction of the impact, whether positive or negative, as well as the magnitude of the impact for individual words. Furthermore, they also compared their explanations to LIME and found that it scored slightly better in terms of overall user preference, though in specific elements, such as positioning of visualization elements, TransSHAP performed slightly better. However, there are limitations in using SHAP due to challenges in feature removal and estimating Shapley values~\cite{zhao2023explainability}. For instance, it is unclear how to select the right baseline for choosing appropriate methods for feature removal. Similarly, calculating Shapley values is computationally expensive, as the complexity grows exponentially with the number of features. Despite these challenges, SHAP is widely used and adapted to LMs, providing sequential visualization explanations for interpreting model predictions.

\subsubsection{Classifier-based Probing}
\label{classifier-based-probing-encoder-only}

Classifier-based probing involves training lightweight classifiers on top of encoder-only models, such as BERT~\cite{mohebbi2021exploring}, to explain the internal representations learned by the model~\cite{koto-etal-2021-discourse}. Unlike feature attribution-based methods, which measure the contribution of input features to a prediction, this approach focuses on understanding what information is encoded within the model’s hidden representations. The process typically begins by freezing the model parameters and generating representations for input words, phrases, or sentences. These representations are then used to train classifiers that identify the linguistic properties or reasoning abilities captured by the model~\cite{zhao2023explainability}. This method is particularly suitable for encoder-only models, as they are pretrained using a masked language modeling (MLM) objective, which enables them to process data bidirectionally and produce strong contextualized word embeddings for each token~\cite{devlin2018bert}. 


To utilize classifier-based probing for interpreting encoder-only models, Tenney et al.~\cite{tenney2019you} introduced edge probing, which investigates how models encode sentence structure across a range of syntactic, semantic, local and long-range phenomena. It focuses on core NLP labeling tasks, such as part-of-speech tagging~\cite{belinkov-etal-2017-evaluating}, semantic
role labeling~\cite{he-etal-2018-jointly}, and coreference resolution~\cite{lee-etal-2018-higher}. Furthermore, this study finds that contextual models like BERT show notable improvements over non-contextual counterparts, particularly on syntactic tasks compared to semantic ones. These findings suggest that contextual embeddings encode syntactic features more strongly than higher-level semantic information. However, a key limitation of edge probing is that it does not investigate the underlying reasons behind layer-wise behavior~\cite{mohebbi2021exploring}, as it does not clarify why certain layers specialize in specific linguistic properties. Additionally, it overlooks the functional role of individual token representations, offering limited insights into how specific tokens contribute to the model’s internal decision-making process.

To address these limitations, Mohebbi et al.~\cite{mohebbi2021exploring} investigate the role of token representations in BERT’s representation space, aiming to explain performance trends across various probing tasks. Their analysis spans a set of surface, syntactic, and semantic tasks, showing that BERT typically encodes the knowledge required for these tasks within specific token representations, particularly in the higher layers. This study finds that most positional information decreases through layers whereas sentence-ending tokens play a partial role in transmitting positional knowledge to higher layers of the model. Additionally, grammatical number and tense information was analyzed throughout the model, observing that BERT encodes verb tense and noun number in the sentence-ending token. However, despite the strong performance of these classifier-based probing techniques, Hewitt and Liang~\cite{hewitt2019designinginterpretingprobescontrol} raise concerns about their faithfulness. Specifically, they argue that high accuracy in probing tasks does not necessarily confirm that the model internally represents the probed features, as the classifier might learn to exploit spurious correlations rather than revealing true model behavior. Consequently, the probing results should be interpreted with caution, as they may not fully capture the underlying mechanisms driving the model’s representations. One effective way to mitigate these concern is through parameter-free probing, which we discuss in the next section.

\subsubsection{Parameter-free probing}

Parameter-free probing techniques avoid training external classifiers or adding additional parameters. Instead, they directly analyze outputs from encoder-only models, such as activation patterns, attention weights, or contextualized embeddings, to infer the presence of linguistic structures~\cite{wu2020perturbed}. This avoids a key pitfall of classifier-based probing, where the probe itself may inadvertently encode task-specific information in its own parameters rather than faithfully revealing the knowledge inherently represented by the model~\cite{hewitt2019designinginterpretingprobescontrol}.

To interpret encoder-only models using a parameter-free probing technique, Wu et al.~\cite{wu2020perturbed} propose perturbed masking, which estimates inter-word correlations and extracts global syntactic information. The method performs a two-stage perturbation: first, a target word is masked; then, an additional word in the sentence is masked to observe how the target’s contextualized representation shifts. This shift quantifies the second word’s influence on the first. From these perturbations, impact matrices are derived from BERT’s outputs, capturing inter-word dependencies similar to attention patterns but without relying on intermediate representations. Using these matrices, syntactic trees are reconstructed, revealing that BERT encodes rich syntactic structures.

Similarly, Zhou et al.~\cite{zhou-srikumar-2021-directprobe} point out that training classifiers as probes can be unreliable because different representations may require different classifiers, and the choice of classifier can significantly influence the estimated quality of a representation. To address this limitation, they propose a parameter-free technique, DIRECTPROBE~\cite{zhou-srikumar-2021-directprobe}, an XAI method designed to interpret and evaluate the quality of representations for NLP tasks without relying on auxiliary classifiers. In this approach, instead of assessing performance using a specific classifier, it approximates the version space that represents the set of all classifiers consistent with the training data by analyzing the geometric structure of the embedding space. Furthermore, through hierarchical clustering, the method partitions the feature space into contiguous regions associated with different labels, providing an interpretable view of how representations organize information. Together, these procedures explain model representations while addressing the challenges posed by training auxiliary classifiers.

\subsubsection{Attention-based Methods}
\label{attention-based-methods-encoder-on}

Encoder-only models such as BERT are built on transformer architectures that employ a fully attention-based mechanism, specifically bidirectional self-attention~\cite{devlin2018bert}. Although previously discussed XAI approaches utilize techniques, such as quantifying each input’s contribution through feature-attribution or training lightweight classifiers to probe what the model has learned, attention-based methods provide a more direct view into model reasoning by visualizing self-attention distributions and showing exactly how weight is allocated across tokens.
To leverage this, various tools have been developed to visualize attention, providing concise summaries of relevant information and facilitating user interaction with LMs~\cite{hoover-etal-2020-exbert}.

To interpret the attention mechanism in encoder-only models, Vig et al.~\cite{vig2019analyzing} introduced BertViz, an open-source tool designed to visualize attention at multiple scales, each offering a distinct perspective on the model’s attention behavior. The tool features a high-level model view, which displays all layers and attention heads within a single interface, and a low-level neuron view, which reveals how individual neurons interact to produce attention patterns. It demonstrates the tool on the BERT model and presents various primary use cases, such as identifying relevant attention heads and linking neurons to model behavior. However, interpreting models solely based on attention weights as faithful explanations can be problematic~\cite{brunner2020identifiabilitytransformers}. Specifically, attention distributions are not guaranteed to reflect the true reasoning process of the model, as different configurations of attention weights can yield similar outputs. Consequently, while BertViz provides valuable visual insights, conclusions drawn from attention patterns alone should be approached with caution, as they may not accurately represent the underlying decision-making mechanisms.

To address these concerns, Hoover et al.~\cite{hoover-etal-2020-exbert} introduced ExBERT, an interactive visualization tool that provides a dynamic and intuitive view of both attention mechanisms and internal representations within transformer models. The tool comprises the attention view, offering an interactive interface allowing users to examine aggregated attention patterns, and the corpus view, presenting aggregate statistics that summarize the hidden representations of a selected token or group of tokens. Using attention visualization and nearest-neighbor search techniques, ExBERT explores what information attention and representations capture, and also helps uncover latent patterns that influence model predictions in text inputs. Despite its various applications, ExBERT is primarily constrained to local analyses, as it focuses on presenting statistics across a few nearest neighbors for one token at a time~\cite{hoover-etal-2020-exbert}. Consequently, it provides limited insights into ability to provide a global understanding of the model’s representational space.

\subsection{XAI Methods for Decoder-only LMs}

XAI methods for decoder-only LMs, such as GPT-4~\cite{openai2023gpt4} and LLaMA-2~\cite{touvron2023llama}, seek to interpret and analyze the reasoning processes of autoregressive architectures that generate text sequentially based on left-to-right context. Unlike XAI methods for encoder-only LMs, which focus on bidirectional self-attention, explainability techniques for decoder-only models account for their unidirectional attention mechanisms~\cite{minaee2025largelanguagemodelssurvey}. Moreover, the massive parameter scales and generative capabilities of these models pose significant challenges for interpretability, often necessitating the use of prompt engineering strategies~\cite{chen2023unleashing}. To address these challenges, various explainability techniques have been proposed, including feature attribution-based methods~\cite{barkan2024llm}, in-context learning (ICL)-based approaches~\cite{brown2020language}, chain-of-thought (CoT) prompting~\cite{zhang2024sea}, and mechanistic interpretability~\cite{meng2022locating}, which will be discussed in the following sections.

\subsubsection{Feature Attribution-based Methods}
\label{feature-attribution-decoder-only}

Feature attribution-based methods aim to quantify how much each individual input token contributes to the model’s final output. As discussed in Section~\ref{feature-attribution-encoder-only}, these methods are well-suited for encoder-only models as they are pretrained using MLM objective and leverage bidirectional self-attention. However, applying them to decoder-only language models poses challenges due to their autoregressive architecture, where token generation is conditioned on the left-hand context. Consequently, recent efforts have focused on designing feature attribution techniques aligned with the autoregressive decoding process inherent in these models.

For instance, to interpret decoder-only models using feature attribution, Barkan et al.~\cite{barkan2024llm} proposed Attributive Masking Learning (AML), a method that explains language model predictions by learning input masks. AML trains an attribution model to identify influential tokens in the input for a given prediction. The optimization follows two complementary objectives: (i) masking as much input data as possible while preserving the model’s original prediction, and (ii) ensuring that applying the complement of the mask significantly alters the prediction. This dual strategy yields explanations aligned with the metric of interest. However, the study has notable limitations. Only one of the five proposed masking approaches, replacement by designated tokens was implemented, leaving alternatives such as replacement by random or contextually predicted tokens unexplored. Moreover, the analysis did not consider broader aspects of explainability, such as stability and fairness.

As another example, Kariyappa et al.~\cite{kariyappa2024progressive} introduced Progressive Inference, a framework for computing input attributions to explain predictions of decoder-only sequence classification models. The core idea is that the classification head can generate intermediate predictions at different points in the input sequence, which, due to the causal attention mechanism, depend only on preceding tokens. This allows obtaining predictions on masked subsequences with negligible computational overhead. Using this insight, the framework proposes two methods: Single Pass-Progressive Inference (SP-PI), which derives attributions from differences between consecutive intermediate predictions, and Multi Pass-Progressive Inference (MP-PI), which leverages Kernel SHAP~\cite{lundberg2017shap} and multiple masked inputs to produce higher-quality attributions. However, this framework has key limitations. For instance, both methods in this framework assume that masked versions of the inputs are valid inputs for the model. Furthermore, MP-PI’s attributions differ from SHAP values because the samples in progressive inference are correlated due to masked attention and intermediate predictions may not accurately represent the model’s prediction on equivalent masked inputs.

\subsubsection{In-Context Learning (ICL)-based Methods}
\label{icl-based-methods}

In-Context Learning (ICL)-based methods explain decoder-only LMs by leveraging their ability to perform tasks directly from examples provided in the input prompt without updating model parameters~\cite{liu2024towards}. Unlike feature attribution-based methods, which explain predictions by quantifying the contribution of individual input tokens, ICL-based methods focus on understanding how models leverage contextual demonstrations to guide their outputs, improving interpretability by making the model’s reasoning process more transparent. This makes them particularly well-suited for decoder-only LMs, as outputs are inherently generated through causal self-attention over left-to-right context, allowing explanations to align with the model’s autoregressive behavior~\cite{zhao2023explainability}.

To enable explainability through in-context learning (ICL) in decoder-only models, Wei et al.~\cite{wei2023larger} interpret how LMs are influenced by semantic priors and input–label mappings. In this study, they explain this behavior of LMs using two approaches. The first approach is flipped-label ICL, where all the exemplar labels are flipped, creating a conflict between semantic prior knowledge and input–label mappings. In this case, although smaller language models tend to ignore in-context flipped labels and rely primarily on semantic knowledge acquired during pretraining, larger models can override these semantic priors when given contradictory exemplars showing emergent abilities. The second approach is semantically unrelated label ICL (SUL-ICL) where the labels are semantically unrelated to the task as semantic cues are removed with model only performing ICL using input-label mappings. In this case, it indicates that the ability to do SUL-ICL also emerges primarily with scale, with sufficiently large models even capable of performing linear classification in such settings. However, this approach has limitations. In particular, while the study provides empirical evidence that larger models exhibit emergent behavior as the model scale increases, the underlying reasons behind the emergence of these complex abilities remain unclear~\cite{chan2022data}. Additionally, this study does not investigate the effect of input distribution.

To address this limitation of not probing input distributions, Li et al.~\cite{li2023saliencyicl} explain the model behavior of decoder-only models under ICL using a fine-grained interpretation of input distributions by editing different components of the input text for task-specific purposes. For example, in sentiment analysis, sentiment-indicative terms in the demonstration inputs are replaced with sentiment-neutral ones, finding that such neutral input perturbations have a smaller impact on model performance compared to changing the ground-truth labels. Furthermore, this approach uses contrastive input–label demonstration pairs constructed through label flipping, input modification, and complementary explanations, and compares their saliency maps using both qualitative and quantitative analyses. With this, it interprets the effects of altering ground-truth labels and adding complementary explanations, providing a better understanding of which parts of the demonstrations contribute most to predictions. However, this study has limitations. In particular, it relies on a relatively small sample size, highlighting the need for larger datasets for more robust evaluation. Additionally, all demonstrations used for ICL are randomly selected, which may influence the model’s reasoning in certain cases.

\subsubsection{Chain-of-Thought (CoT) Prompting}
\label{cot-prompting}

Chain-of-Thought (CoT) prompting enables decoder-only language models to articulate intermediate reasoning steps before producing a final answer~\cite{wei2022chain}. In contrast to In-Context Learning (ICL)-based methods, which rely on demonstrations to learn task-specific patterns without exposing the underlying reasoning, CoT prompting encourages models to generate step-by-step rationales in natural language~\cite{wang2022self}. 
This technique aligns well with the autoregressive nature of decoder-only LMs, which facilitates the generation of coherent reasoning chains alongside answers. As a result, CoT prompting not only enhances performance on tasks requiring multi-step reasoning but also improves interpretability by providing insight into the model’s decision-making process~\cite{zhao2023explainability}.

To enhance explainability in decoder-only LMs, Zhou et al.~\cite{zhou2022least} proposed Least-to-Most prompting, a CoT-based strategy designed to make the model’s intermediate reasoning steps more interpretable while improving performance. The approach operates in two stages: first, it decomposes a complex problem into a sequence of simpler subproblems; second, it solves these subproblems sequentially, where each solution builds on the outcomes of the previous ones. Both stages are implemented via few-shot prompting, requiring no additional training or fine-tuning. By explicitly structuring the reasoning process, this method improves interpretability while achieving strong performance on tasks involving symbolic manipulation, compositional generalization, and mathematical reasoning. However, this approach has limitations. Its explainability is constrained by poor cross-domain generalization since decomposition prompts designed for one domain (\textit{e.g.}, mathematical word problems) are often ineffective for others (\textit{e.g.}, commonsense reasoning). As a result, maintaining transparency and performance requires domain-specific prompt engineering.

Similarly, another Chain-of-Thought prompting technique is proposed by Jie et al.~\cite{wei-jie-etal-2024-interpretable}, presenting a multifaceted evaluation of interpretability, examining dimensions such as faithfulness, robustness, and utility across multiple commonsense reasoning benchmarks. Unlike studies restricted to a single prompting strategy, this work investigates a broad spectrum of widely used prompting techniques, enabling a more exhaustive assessment. Furthermore, a simple explainability alignment method, termed Self-Entailment Alignment Chain-of-Thought (SEA-CoT), is introduced to enhance interpretability within the reasoning chain. This method operates similarly to Self-Consistency~\cite{wang2022self} but additionally enforces consistency between the intermediate reasoning steps and the supporting context. However, despite its utility for enhancing explainability, this study does not explore techniques that ground the model’s responses using external knowledge, leaving open issues related to self-hallucination, where language models produce plausible but non-factual content. Moreover, it does not investigate hybrid approaches, such as neuro-symbolic AI~\cite{colelough2025neuro}, which combine the learning capabilities of neural networks with the decision-making frameworks of symbolic systems which are inherently interpretable.

\subsubsection{Mechanistic Interpretability}
\label{mechanistic-interpretability}

Mechanistic interpretability aims to understand the internal computations of decoder-only LMs by reverse-engineering how their parameters, neurons, and attention heads collectively contribute to generating predictions~\cite{olah2020zoom}. In contrast to interpretability approaches such as CoT prompting, which infer reasoning patterns from observable outputs, mechanistic interpretability focuses on uncovering how models internally represent, manipulate, and propagate information during inference~\cite{nanda2023progress}. This approach is particularly well-suited for decoder-only LMs due to their autoregressive architecture with causal self-attention, which enables a structured flow of information from left to right, allowing researchers to trace how previous tokens influence subsequent predictions~\cite{minaee2025largelanguagemodelssurvey}.

To interpret decoder-only LMs using mechanistic interpretability, Meng et al.~\cite{meng2022locating} analyze how factual associations are stored and recalled in autoregressive transformer models. They identify localized, directly editable computations within mid-layer feed-forward modules that mediate factual predictions. Using a causal intervention, they pinpoint neuron activations critical for these predictions and introduce Rank-One Model Editing (ROME)~\cite{meng2022locating} to modify specific associations. The results show that ROME proves effective on zero-shot relation extraction tasks and a challenging counterfactual dataset, maintaining both specificity and generalization. These findings highlight the central role of mid-layer feed-forward modules in storing factual knowledge and demonstrate the feasibility of directly manipulating computational mechanisms for model editing. However, this approach has limitations. For instance, ROME and causal tracing primarily focus on factual associations, leaving other forms of learned knowledge, such as logical, spatial, and numerical reasoning, largely unexplored. Additionally, despite successful modification of factual associations, models may still generate plausible yet unsupported facts, which constrains their reliability as sources of accurate information.

On the other hand, Conmy et al.~\cite{conmy2023towards} automate a key step in mechanistic interpretability by identifying connections between abstract neural units that form a circuit. Specifically, they locate a subgraph of the model responsible for a target behavior. To achieve this, they introduce Automatic Circuit DisCovery (ACDC)~\cite{brunner2020identifiabilitytransformers}, an algorithm that emulates researchers’ circuit-identification workflows by adapting Subnetwork Probing~\cite{cao-etal-2021-low} and Head Importance Score for Pruning~\cite{NEURIPS2019_2c601ad9}. The study validates ACDC by reproducing prior results, such as successfully rediscovering all five component types in a circuit in GPT-2 Small~\cite{brown2020language} computing the Greater-Than operation. However, ACDC has certain limitations, as it struggles to fully automate activation patching. In particular, it may overlook certain classes of abstract units that are part of the circuit, such as the negative name mover heads from IOI~\cite{wang2023interpretability}. Moreover, the algorithm’s behavior is highly sensitive to hyperparameter configurations and metric selection, which can result in inconsistent and non-robust performance across different settings.

\sloppy
\subsection{XAI Methods for Encoder-decoder LMs}

XAI methods for encoder-decoder LMs, such as T5~\cite{raffel2020exploring} and BART~\cite{lewis2019bartdenoisingsequencetosequencepretraining}, seek to explain the internal decision-making processes of models with a dual architecture that combines both an encoder and a decoder component~\cite{wang2025history}. Unlike XAI techniques for encoder-only LMs, which primarily focus on interpreting bidirectional self-attention, and decoder-only LMs, which emphasize causal self-attention for autoregressive generation, explainability methods for encoder-decoder architectures emphasize cross-attention mechanisms that transfer information between the encoder and decoder, as well as the sequence-to-sequence objectives on which these models are pretrained~\cite{minaee2025largelanguagemodelssurvey}. To interpret these models, various explainable AI (XAI) techniques have been proposed, including feature attribution–based methods~\cite{barkan2024llm}, classifier-based probing~\cite{koto-etal-2021-discourse}, attention-based methods~\cite{juraska2021attention}, and self-explanation methods~\cite{yordanov2022few}, as discussed below.


\sloppy
\subsubsection{Feature Attribution-based Methods}
\label{feature-attribution-encoder-decoder}

Feature attribution-based methods, as discussed in Sections~\ref{feature-attribution-encoder-only} and~\ref{feature-attribution-decoder-only}, aim to quantify the contribution of individual input tokens or features to a model’s predictions~\cite{lundberg2017shap}. In contrast to their application in encoder-only and decoder-only LMs, these methods for encoder-decoder models consider both the bidirectional contextual representations generated by the encoder and the auto-regressive decoding process in the decoder~\cite{raffel2020exploring}. Consequently, feature importance is influenced not only by the relationships among the input tokens but also by the cross-attention mechanisms that facilitate information transfer between the encoder and decoder~\cite{enouen-etal-2024-textgenshap}. To address these complexities, researchers have proposed various approaches tailored to sequence-to-sequence models while considering their dual architecture.

For example, to apply feature attribution-based methods for XAI in encoder-decoder models, Liu et al.~\cite{liu2024reliabilityexplainabilitylanguagemodels} conduct an empirical study on various encoder-decoder LMs, including CodeT5~\cite{wang-etal-2023-codet5}, CodeReviewer~\cite{li2022automatingcodereviewactivities}, and CodeT5+~\cite{wang-etal-2023-codet5}, to evaluate the explainability of automated program generation approaches. This study employs gradient-based SHAP~\cite{lundberg2017shap} to examine and interpret the models’ decision-making processes by highlighting tokens that significantly contribute to code transformation. The analysis demonstrates that these models often overlook critical tokens and exhibit limited robustness. Furthermore, the findings reveal that, while the models can effectively recognize code grammar and structural patterns, their performance remains sensitive to changes in input sequences. However, this study has certain limitations. For instance, the evaluation was conducted using different program generation models with their default architectures and hyperparameter settings as reported in the original papers. While this ensures consistency across models when analyzing their explainability, it may not reflect their optimal performance, as hyperparameter tuning could potentially yield better results.

Similarly, another XAI technique for encoder-decoder LMs, proposed by Enouen et al.~\cite{enouen-etal-2024-textgenshap}, is TextGenSHAP, this approach extends Shapley values for text generation while maintaining computational efficiency for large-scale encoder-decoder models, such as T5-large~\cite{chung2022scalinginstructionfinetunedlanguagemodels} and T5-XXL~\cite{chung2022scalinginstructionfinetunedlanguagemodels}. Specifically, it focuses on explaining open-ended text generation using long prompts, particularly for abstractive question answering from retrieval-augmented documents. Moreover, TextGenSHAP achieves notable runtime improvements for token- and document-level explanations and further enhances retrieval-augmented generation by localizing important words and re-ranking retrieved passages. Furthermore, experiments on open-domain QA benchmarks demonstrate that TextGenSHAP significantly improves both retrieval recall and prediction accuracy. However, despite these improvements, the approach has certain limitations. In particular, generating explanations can still be computationally expensive for many real-world applications, making integration challenging. Furthermore, the study does not explore scenarios involving models distributed across multiple GPUs or TPUs, meaning it cannot guarantee speedups under hardware-specific optimizations.

\subsubsection{Classifier-based Probing}

Classifier-based probing, as discussed in Section~\ref{classifier-based-probing-encoder-only}, has proven effective for interpreting encoder-only language models by training shallow classifiers on frozen model representations to reveal model behavior and prediction logic~\cite{mohebbi2021exploring}. This approach can be extended to encoder-decoder models, where probing is applied across the sequence-to-sequence architecture involving both the encoder and decoder. However, the added complexity of encoder-decoder models—particularly the interplay introduced by components such as cross-attention—poses significant challenges for interpretability. These interactions complicate the analysis of internal representations and obscure how predictions are formed~\cite{raffel2020exploring}.

To apply classifier-based probing to encoder–decoder LMs, Koto et al.~\cite{koto-etal-2021-discourse} introduced document-level discourse probing to interpret the ability of models such as T5~\cite{raffel2020exploring} and BART~\cite{lewis2019bartdenoisingsequencetosequencepretraining} to capture discourse-level relations. This approach helps identify whether, and at which layers, the model captures discourse structure. The study shows that BART’s encoder layers perform best in capturing discourse information for probing tasks such as discourse connective~\cite{nie-etal-2019-dissent}, RST nuclearity~\cite{mann-thompson-1986-assertions} and RST relation~\cite{mann-thompson-1986-assertions} while the decoder performs comparatively poorly. This explains why decoder layers are less effective for language understanding than encoder layers and instead focus more on sequence generation. Similar trends were observed in the T5 encoder–decoder model, which is also trained as a denoising autoencoder. Moreover, this study demonstrates that discourse knowledge is unevenly distributed across model layers, influenced by their pre-training objective and model architecture.

On the other hand, Li et al.~\cite{li-etal-2021-implicit} utilize classifier-based probing to investigate whether the effectiveness of encoder-decoder LMs such as T5~\cite{raffel2020exploring} and BART~\cite{lewis2019bartdenoisingsequencetosequencepretraining} arises solely from capturing surface-level word co-occurrence patterns or whether these models also represent and reason about the described world. They identify contextual word representations that function as models of entities and situations as they evolve within a discourse. These representations exhibit similarities to dynamic semantic models, as they enable a linear readout of an entity’s properties and relations and can be manipulated with predictable effects on language generation. This study suggests that encoder-decoder models rely, at least in part, on dynamic representations of meaning and implicit simulation of entity states, even when trained solely on textual data. However, it has limitations. For instance, the LM outputs and implicit state representations are far from perfect; even in relatively simple tasks, the best-case scenario recovers complete information states correctly only about half of the time. This indicates that the semantic representations lack sufficient expressive power and reliability, falling short of what is required for a human-like explanation.

\subsubsection{Attention-based Methods}
\label{attention-based-methods-encoder-decoder}


Attention-based methods, as discussed in Section~\ref{attention-based-methods-encoder-on}, aim to interpret model behavior by analyzing the distribution of attention weights across tokens~\cite{raffel2020exploring}. While attention mechanisms have
been widely used for interpretability in encoder-only LMs, which rely on bidirectional self-attention, encoder-decoder architectures introduce additional complexity due to their dual structure comprising encoder and decoder components linked through cross-attention~\cite{minaee2025largelanguagemodelssurvey}. This architectural design makes it challenging to attribute model decisions and to understand how information flows across different layers.

To interpret the inner workings of encoder-decoder LMs using attention-head methods, Juraska et al.~\cite{juraska2021attention} propose a semantic attention-guided decoding method (SEAGUIDE) which visualizes the attention weight distribution across individual cross-attention layers in the decoder for different inputs, revealing multiple universal patterns that can be used to determine whether input slots are mentioned or omitted in the generated output. Specifically, the first pattern appears in the lowest attention layers, whose role is to track the corresponding input token associated with the token generated by the decoder in the preceding step. This pattern, called the verbatim slot mention pattern, can be captured by maximizing the weight across all attention heads in the first layer of the decoder. On the other hand, paraphrased slot mention patterns are captured in the higher layers when a corresponding token is about to be generated next. In these higher layers, the cross-attention weights gradually shift toward input tokens corresponding to information most likely to appear next in the output, capturing increasingly abstract concepts. 
Furthermore, the unrealized slot mention pattern is captured by averaging the attention weights over all layers at the final decoding step, which helps to reduce any undesired side effects of identifying paraphrased slot mention patterns. 
However, SEAGUIDE’s ability to detect slot-related errors is limited, as it focuses primarily on missing and incorrect slot mentions. It also struggles to reliably identify duplicate slots because the decoder inherently attends to certain input tokens across multiple non-consecutive steps.

Another attention-based method for explaining encoder-decoder models is proposed by Cao et al.~\cite{cao2021attention}, which introduces an inference-time attention head masking mechanism to interpret encoder-decoder attentions in abstractive summarization. To explain the content selection behavior of individual heads, the study investigates leveraging encoder-decoder attentions to guide content selection and enhance summary informativeness. Specifically, it applies attention head masking based on oracle content selection labels, constructed by aligning the reference summary with the source article and iteratively identifying the longest common subsequences between them. The findings show that applying oracle masking to the top attention layers yields the greatest improvement in summary informativeness. The study also explores whether masking multiple heads enhances content selection and whether these heads act synergistically. By progressively masking an increasing number of heads at each layer, it finds that the largest performance gain occurs when masking 15 out of 16 heads in the top layer, indicating strong collaboration among multiple heads for content selection. Furthermore, the study analyzes the types of words attention heads focus on: for each generated token during decoding, the input token with the highest attention weight (the attendee) is categorized into groups such as SALIENT, CONTENT, and FIRST/LAST tokens. The results reveal that higher layers tend to focus on tokens directly copied into the output, whereas lower layers attend more to salient words that are excluded from the current generation. However, this approach has limitations. For tasks that demand a comprehensive understanding of the source such as XSum~\cite{narayan-etal-2018-dont} the method may falter. Specifically masking ``unimportant'' tokens can inadvertently strip away essential contextual information~\cite{ma2021global}, leading to unstable performance and diminished effectiveness.

\subsubsection{Self-explanation Methods}

Self-explanation-based methods aim to make encoder-decoder LMs more interpretable by generating human-readable justifications alongside task predictions. Unlike attention-based approaches that rely on interpreting internal attention weights to infer model reasoning~\cite{juraska2021attention, cao2021attention}, these methods produce explanations as part of the output sequence, leveraging the generative capacity of the decoder~\cite{narang2020wt5}. This aligns with encoder-decoder architectures, where explanation and prediction can be treated as a unified text generation task~\cite{rajani2019explain}. It also enhances transparency by directly linking the model’s decision-making process with the generated explanations, enabling users to better understand and evaluate the rationale behind the predictions~\cite{narang2020wt5, yordanov2022few}.

A popular approach to implement self-explanations for encoder-decoder models is introduced by Narang et al.~\cite{narang2020wt5} through WT5, which leverages the text-to-text framework to train models to generate natural language explanations alongside their predictions. This method does not require any modifications to the loss function, training objectives, or decoding procedures; instead, the model is simply trained to produce an explanation after generating the predicted output. The experimental results of this technique demonstrate that it achieves state-of-the-art performance on explainability benchmarks while also enabling learning from a limited number of annotated explanations and facilitating the transfer of rationalization capabilities across different datasets. However, a key limitation of this approach is that it typically requires large amounts of annotated explanations during training, which can be costly and time-consuming to obtain. For many real-world applications, this reliance on extensive data becomes a significant bottleneck, limiting the scalability and practical deployment of such methods.

Extending the paradigm of self-explanation techniques for encoder-decoder LMs, Yordanov et al.~\cite{yordanov2022few} propose three few-shot transfer learning approaches for generating natural language explanations (NLEs) and incorporate an additional method adapted from Erliksson et al.~\cite{10.1007/978-3-030-80599-9_8} to enhance computational efficiency. These four approaches integrate multi-task learning with fine-tuning to enable knowledge transfer from a parent task, which has abundant training labels, to a child task that has limited NLE annotations but sufficient labels. Specifically, the framework transfers explanatory capabilities from a large-scale natural language inference dataset (e-SNLI)~\cite{10.5555/3327546.3327624} to two downstream child tasks: (1) complex pronoun resolution cases, for which the authors develop the small-e-WinoGrande dataset of NLEs based on WinoGrande~\cite{Sakaguchi_LeBras_Bhagavatula_Choi_2020}, and (2) commonsense validation (ComVE)~\cite{wang-etal-2020-semeval}. Moreover, the results highlight the potential of few-shot out-of-domain transfer learning for NLEs and provide insights into the effectiveness of different learning strategies in this context. However, this approach has certain limitations. Although the proposed training methods can be applied to languages other than English, they require the availability of a large parent dataset containing NLE annotations. Additionally, achieving effective performance may also depend on the existence of a high-quality pre-trained generative language model for the target language.

\section{Evaluation of Explanations}

After discussing various XAI methods within a systematic taxonomy based on the transformer architectures of LMs, it is essential to evaluate the quality and reliability of the explanations generated by these approaches. These explanations are typically assessed not only for their understandability to humans but also for how accurately they reflect the model’s underlying reasoning process~\cite{10.1145/3583558}.
This dual requirement has led to two complementary evaluation dimensions: plausibility and faithfulness~\cite{zhao2023explainability}. Specifically, plausibility measures the extent to which explanations align with human intuition and appear reasonable, whereas faithfulness examines whether explanations reliably capture the model’s true decision-making process~\cite{pmlr-v235-chen24bl}. Together, these dimensions provide a structured framework for evaluating and comparing interpretability methods across encoder-only, decoder-only, and encoder-decoder architectures.

\subsection{Plausibility}
\label{subsec:plausibility}

Plausibility refers to the extent to which an explanation is perceived as reasonable or convincing by humans~\cite{jacovi-goldberg-2020-towards}. To evaluate explanations using plausibility, researchers have proposed several methodologies:

\begin{itemize}
    \item DeYoung et al.~\cite{deyoung2020eraser} introduce the Evaluating Rationales and Simple English Reasoning (ERASER), which measures how closely model-generated explanations align with human-annotated rationales across multiple datasets. This technique incorporates two types of metrics depending on whether models perform discrete or soft selection.

    \bigskip

In the discrete setting, it employs the Intersection-over-Union (IoU) metric~\cite{everingham2010pascal}, which permits credit assignment for partial matches between predicted rationale span and human-annotated rationale span. Here, the predicted rationale span refers to a contiguous segment of input text highlighted by the model as its explanation, while the human-annotated rationale span denotes the corresponding gold evidence marked by annotators. Formally, IoU is defined as: 
    
    \begin{equation}
    \mathrm{IoU} = 
    \frac{\lvert B_p \cap B_{gt} \rvert}
         {\lvert B_p \cup B_{gt} \rvert}
    \end{equation}
    
    where $B_p$ denotes the predicted rationale span and $B_{gt}$ denotes the human-annotated ground-truth rationale span.

    \bigskip

    A prediction is considered a match when the overlap between the predicted rationale and a ground-truth rationale exceeds $50\%$, \textit{i.e.}, $\mathrm{IoU} > 0.5$. This threshold ensures that only predicted spans sufficiently aligned with human-annotated rationales are counted as correct. 
    These matched predictions are then used to compute alignment metrics such as F1-score, which  measure  how well the model’s explanations agree with human rationales.
            
    \bigskip

    In the soft selection setting, ERASER evaluates token rankings by rewarding models for assigning higher importance scores to tokens annotated by humans. Specifically, it calculates the Area Under the Precision–Recall Curve (AUPRC)~\cite{deyoung2020eraser}, obtained by sweeping a threshold over token-level scores:
    
    \begin{equation}
    \mathrm{AUPRC} = \sum_{n=1}^{N} \left( R_n - R_{n-1} \right) P_n,
    \end{equation}

    \bigskip
    
    where $P_n$ and $R_n$ denote precision and recall at the $n$-th threshold, 
    respectively.

    \bigskip

    Here, AUPRC quantifies how well a model ranks human-annotated rationale tokens higher than irrelevant tokens. A higher AUPRC indicates that the model consistently assigns greater importance to human-labeled rationales, thus exhibiting better plausibility.

\item Ding and Koehn~\cite{ding-koehn-2021-evaluating} propose a test to evaluate the plausibility of explanations by leveraging lexical agreement annotations as a form of ground truth. They divide the nouns in the prefix into two distinct sets: the cue set $C$, which contains nouns sharing the same morphological number as the verb in the sentence, and the attractor set $A$, which contains nouns with a different morphological number from the verb in the sentence. Formally, let $\psi(w)$ denote the saliency score assigned to a word $w$ by the explanation method. Then, based on the prediction $y$ made by the model $M$, the test is conducted under one of the following two scenarios:
\bigskip

Expected: When $y$ corresponds to the verb with the correct number, the interpretation passes the test if 
   \begin{equation}
    \max_{w \in C} \psi(w) > \max_{w \in A} \psi(w)
   \end{equation}

   In this case, a plausible explanation should assign higher importance to cue words than to attractors, indicating that the model relied on the correct grammatical cues when making the prediction.

\bigskip

     Alternative: When $y$ corresponds to the verb with the incorrect number, the interpretation passes the test if 
     \begin{equation}
    \max_{w \in C} \psi(w) < \max_{w \in A} \psi(w)
     \end{equation}

Here, the explanation is considered plausible if the attractor words receive higher importance than the cue words, showing that the model was misled by the wrong grammatical cues.

   \item Shen et al.~\cite{shen2022interpretability} evaluate 
plausibility based on how well the rationales provided by the model agree with human-annotated ones. To quantify this, the approach adopts the token-level F1-score, computed from the overlap between predicted and annotated rationale tokens. Since an instance may contain multiple golden rationale sets, the set achieving the highest F1-score with the predicted rationale is selected as the ground truth for the current prediction:

\begin{equation}
\text{Token-F1} = 
\frac{1}{N} \sum_{i=1}^{N} 
\left( 
\frac{2 \times P_i \times R_i}{P_i + R_i} 
\right),
\quad 
P_i = \dfrac{|S^p_i \cap S^g_i|}{|S^p_i|},
\quad 
R_i = \dfrac{|S^p_i \cap S^g_i|}{|S^g_i|}
\end{equation}

\bigskip

\noindent
where $P_i$ is the precision, which measures the proportion of predicted rationale tokens that overlap with the human rationale, $R_i$ is the recall, which measures the proportion of human rationale tokens that overlap with the predicted rationale, $N$ denotes the total number of test instances, $S^p_i$ and $S^g_i$ represent the rationale sets of the $i$-th instance provided by the model and the human, respectively.

\bigskip

\noindent
A higher Token-F1 indicates better alignment between model-predicted and human-annotated rationales, reflecting greater plausibility, while a lower score suggests less interpretable and less human-aligned explanations.

\end{itemize}

The techniques discussed above help evaluate the plausibility of explanations produced by various XAI methods across different categories. However, several studies emphasize that evaluating the plausibility of explanations is fundamentally different from assessing their correctness, and the two criteria should be treated separately~\cite{10.1145/3583558}. For example, Jacovi et al.~\cite{jacovi-goldberg-2020-towards} argue that explanations perceived as plausible do not necessarily reflect the model’s true reasoning process. Similarly, Petsiuk et al.~\cite{petsiuk2018rise} suggest that excluding humans from the evaluation loop can lead to more objective assessments by focusing on the classifier’s decision-making process rather than subjective judgments. Such approaches reduce potential human bias while saving time and computational resources. Furthermore, Gilpin et al.~\cite{8631448} highlight that explanations perceived as unreasonable may indicate either limitations in the model’s ability to process information in a reasonable way or shortcomings in the explanation generation method itself.

\subsection{Faithfulness}

While plausibility-based evaluations focus on whether explanations appear reasonable to humans, such rationales may not necessarily correspond to the model’s true decision-making process~\cite{jacovi-goldberg-2020-towards}. In contrast, faithfulness refers to the extent to which an explanation accurately captures the internal reasoning of the model, ensuring that the identified rationale is causally linked to and meaningfully contributes to the model’s prediction~\cite{mathew2021hatexplain}. To assess the faithfulness of explanations, researchers have introduced various evaluation methodologies:

\begin{itemize}

     \item DeYoung et al.~\cite{deyoung2020eraser} introduce two widely adopted metrics for evaluating faithfulness: comprehensiveness and sufficiency. Firstly, comprehensiveness measures whether the selected rationales include all features necessary for the model’s prediction. The comprehensiveness score is defined as:

    \begin{equation}
    \text{Comprehensiveness} = m(x_i)_j - m(\tilde{x}_i)_j
    \end{equation}

    \bigskip

    \noindent
    where $x_i$ denotes the input instance, $\tilde{x}_i$ denotes the contrast example obtained by removing predicted rationale $r_i$ from $x_i$, and  $m(\cdot)_j$ denotes the model’s predicted probability for the predicted class $j$ in a classification setting.
   
    \bigskip

    A higher score indicates greater faithfulness, as it implies that removing the rationales significantly decreases the model’s confidence. Conversely, a low score suggests that the rationales contributed little to the decision, and a negative score indicates an even less faithful scenario where the model becomes more confident after removing the rationales.
    
    \bigskip

Secondly, sufficiency, in contrast, measures whether the extracted rationales alone are adequate for the model to reproduce its original decision. Formally, it is defined as:
    
    \begin{equation}
    \text{Sufficiency} = m(x_i)_j - m(r_i)_j
    \end{equation}

    \bigskip

     \noindent
    where $x_i$ denotes the input instance, $r_i$ denotes the predicted rationale, and  $m(\cdot)_j$ denotes the model’s predicted probability for the predicted class $j$ in a classification setting.

    \bigskip
    
    Lower sufficiency scores imply higher faithfulness, as they indicate that $r_i$ alone contains enough information to make the same prediction with minimal confidence loss. In contrast, higher scores suggest that the rationales are insufficient and the model relies on additional context from the rest of the input.

    \item Ding and Koehn~\cite{ding-koehn-2021-evaluating} propose consistency tests to evaluate the faithfulness of explanations, focusing on two aspects: model consistency, which measures the stability of feature importance scores under changes in model configuration, and input consistency, which measures the stability of scores under perturbations to the input.

\bigskip

To evaluate model consistency, the similarity between feature importance scores of the original model and a smaller distilled model trained to mimic it is computed. Although the original model and the distilled model differ in configuration, the distillation process encourages the smaller model to preserve the decision behavior of the original. This setup provides a meaningful test of whether explanations truly capture the model’s internal reasoning. Formally, model consistency is defined as:


\begin{equation}
\text{ModelConsistency}(M, M', F)
= \rho \left( \Psi_{M}(F), \Psi_{M'}(F) \right),
\end{equation}

\bigskip

where $M$ denotes the original model, $M'$ is the smaller distilled model trained to mimic $M$, $\Psi_{M}(F)$ and $\Psi_{M'}(F)$ represent the feature importance scores, $F$ denotes the input features, and $\rho(\cdot,\cdot)$ is the Pearson correlation coefficient.

\bigskip


Here, higher model consistency indicates that the saliency method produces stable and faithful explanations across models with similar decision behavior, whereas a low model consistency suggests sensitivity to configuration changes, undermining faithfulness.

\bigskip

To evaluate input consistency, tokens in the input are substituted while preserving the underlying decision, and the similarity between feature importance scores of the original and perturbed inputs is measured. Formally, input consistency is defined as:

\begin{equation}
\text{Input Consistency}(F, F')
= \rho \left( \Psi(F), \Psi(F') \right),
\end{equation}

\bigskip
where $F$ and $F'$ denote the original and perturbed input features, respectively; $\Psi(F)$ and $\Psi(F')$ denote their corresponding feature importance scores; and $\rho(\cdot,\cdot)$ represents the Pearson correlation coefficient.

\bigskip

  Here, higher input consistency indicates that the feature importance remains stable under controlled changes in input, thus signifying greater faithfulness of the interpretation. Conversely, a lower input consistency suggests that the attribution is sensitive to irrelevant variations, indicating reduced faithfulness.

\item Shen et al.~\cite{shen2022interpretability} evaluate the faithfulness of language models by measuring the consistency of rationales under input perturbations by adopting Mean Average Precision (MAP)~\cite{wang2021dutrust}. This metric quantifies the order consistency between two ranked token lists: the rationale extracted from the original input and that from the perturbed input. Formally, MAP is defined as:

\begin{equation}
\text{MAP} =
\frac{1}{|X^{p}|}
\sum_{i=1}^{|X^{p}|}
\frac{\sum_{j=1}^{i} G(x_{j}^{p}, X_{1:i}^{o})}{i}
\end{equation}

\bigskip

where $X^{o}$ and $X^{p}$ denote the ranked token lists obtained from the original and perturbed inputs, respectively, and $|X^{p}|$ is the number of tokens in $X^{p}$. The subset $X^{o}_{1:i}$ contains the top-$i$ tokens from $X^{o}$. The indicator function $G(x, Y)$ returns $1$ if $x \in Y$ and $0$ otherwise.

\bigskip

A higher MAP score indicates that the ranking of important tokens remains consistent under perturbations, demonstrating stronger faithfulness of the rationale. In contrast, a lower MAP score reflects weaker faithfulness, suggesting that the rationale does not reliably capture the model’s reasoning.

    \item Chrysostomou et al.~\cite{chrysostomou-aletras-2021-improving} evaluate faithfulness by measuring the impact of removing the most relevant tokens identified by explanations. Specifically, they introduce two metrics: Decision Flip – Most Informative Token and Decision Flip – Fraction of Tokens.

\bigskip

Firstly, Decision Flip – Most Informative Token measures the percentage of predictions that change when the single most important token is removed. Formally, it is defined as:

\begin{equation}
\text{DF}_{\text{MIT}} = \frac{1}{N} \sum_{i=1}^{N} \mathbf{1}\!\left[ m(x_i) \neq m(x_i \setminus t_i^{*}) \right]
\end{equation}

where $N$ is the total number of test instances, $m(\cdot)$ denotes the model prediction, $x_i$ is the input instance, $t_i^{*}$ is the most important token according to the explanation method, and $\mathbf{1}[\cdot]$ is the indicator function. 

\bigskip

Here, a higher score indicates stronger faithfulness, since removing the most important token is sufficient to change the model’s prediction. Conversely, a lower score suggests that the top-ranked token was not truly decisive for the model’s decision.

\bigskip

Secondly, Decision Flip – Fraction of Tokens computes the average proportion of tokens that must be removed to induce a change in the model’s prediction. Formally, it is defined as:

\begin{equation}
\text{DF}_{\text{Frac}} = \frac{1}{N} \sum_{i=1}^{N} \frac{|R_i|}{|x_i|}
\end{equation}

where $R_i$ is the minimal set of top-ranked tokens that need to be removed from $x_i$ to flip the prediction, $|x_i|$ is the sequence length, and $N$ is the total number of test instances.

\bigskip

Here, a lower score indicates stronger faithfulness, as only a small proportion of top-ranked tokens need to be removed to alter the prediction. Conversely, a higher score means that many tokens must be removed, suggesting the explanation is less focused on the truly influential evidence.

\end{itemize}

The approaches discussed above are employed to assess the faithfulness of explanations across different explainability methods. Moreover, faithfulness thus complements plausibility by ensuring that explanations are not only persuasive to humans but also accurately reflect the model’s internal reasoning. Achieving a balance between these perspectives is crucial for developing explanations that are both trustworthy and interpretable.

\section{Challenges and Future Directions}
\label{sec:challenges-future}

Despite significant progress in developing XAI techniques for LMs, several open challenges remain. This section discusses these challenges and outlines potential avenues for future research, focusing on key aspects such as the scalability of explanations, the need for methods that operate without ground-truth rationales, the trade-offs between performance and interpretability, the explainability of multimodal systems, and the broader ethical and societal implications. The following subsections examine each of these challenges and highlight promising directions for addressing them.

\subsection{Scalability of Explanations}
As language models continue to grow in size and complexity, generating faithful and interpretable explanations becomes increasingly challenging. Many existing XAI techniques, such as attribution-based methods~\cite{lundberg2017shap}, attention visualizations~\cite{juraska2021attention}, and mechanistic interpretability~\cite{brunner2020identifiabilitytransformers}, require extensive computations or repeated model evaluations, leading to substantial resource demands. These challenges are further amplified in large-scale models with hundreds of billions of parameters, where token-level or layer-wise explanations often become impractical for real-time or resource-constrained applications. Future research should explore strategies that improve the efficiency of explanation techniques while maintaining their interpretability, enabling scalable and practical solutions for modern LMs.

\subsection{Explanation without Ground Truths}

Ground-truth explanations for language models are often inaccessible, making it challenging to evaluate the quality and reliability of generated explanations~\cite{zhao2023explainability}. While existing benchmarks, such as ERASER~\cite{deyoung2020eraser}, provide human-annotated rationales for plausibility and faithfulness assessment, their coverage is limited in terms of tasks and domains. Moreover, for many open-ended tasks such as summarization, dialogue generation, or creative writing, multiple valid explanations may exist, and identifying a single correct rationale becomes inherently ambiguous~\cite{10.1145/3583558}. This lack of standardized ground-truth references highlights the need for alternative evaluation strategies, such as those focusing on internal model consistency, causal alignment, and cross-metric validation, rather than relying exclusively on annotated datasets.

\subsection{Performance vs. Interpretability}

Recent advances in XAI techniques such as chain-of-thought prompting, feature attribution methods, and self-explanation strategies have enhanced the interpretability of language models by providing greater transparency into their decision-making processes~\cite{liu2024reliabilityexplainabilitylanguagemodels, yordanov2022few,zhou2022least}. However, these techniques often introduce additional computational overhead and, in some cases, lead to reductions in task performance~\cite{crook2023revisiting}. Conversely, models optimized exclusively for predictive accuracy may yield limited or no meaningful explanations, constraining their applicability in high-stakes domains where accountability and transparency are critical. Addressing this problem requires research efforts aimed at developing approaches that balance predictive performance with interpretability, ensuring that explanations remain reliable without significantly compromising model effectiveness.

\subsection{Explainibility of Multimodal models}
Recent advances in multimodal models, which integrate information from diverse modalities such as text, images, audio, and video, have expanded the capabilities of language models but also introduced new challenges for explainability~\cite{10689601}. Unlike unimodal LMs, where explanations can focus solely on textual inputs and outputs, multimodal systems require understanding how information is combined, aligned, and propagated across different modalities. Existing XAI techniques, which are primarily designed for text-only models, often fail to capture these cross-modal interactions and dependencies~\cite{MERSHA2024128111}. Addressing this challenge requires XAI frameworks that jointly analyze contributions from multiple modalities, enabling interpretability while preserving the distinct roles of each modality in the decision-making process.

\subsection{Ethical and Societal Concerns}

As language models are increasingly deployed in high-stakes domains such as healthcare, finance, education, and law, their explanations carry significant ethical and societal implications~\cite{lee2020biobert,chen2021evaluating,ng2024educational}. Explanations that are incomplete, misleading, or biased can undermine user trust and lead to harmful outcomes. Moreover, the absence of standardized explainability practices complicates regulatory compliance and accountability, especially when decisions directly affect individuals or communities. Addressing this challenge requires developing explanation frameworks that emphasize fairness, transparency, and inclusivity while mitigating biases in both models and their generated explanations~\cite{zhao2023explainability}. Furthermore, explainability methods should consider diverse user needs, ensuring that explanations remain accessible and meaningful across different contexts and populations.

\section{Conclusion}
\label{sec:conclusion}

In this survey, we provided a comprehensive review of XAI techniques for language models, systematically organizing existing methods according to their underlying transformer architectures: encoder-only, decoder-only, and encoder-decoder. By analyzing a wide range of approaches, we highlight how architectural differences shape the design and effectiveness of explainability methods by examining their strengths and limitations. We also discussed evaluation strategies through the complementary lenses of plausibility and faithfulness, emphasizing the need for explanations that are both interpretable to humans and faithful to model reasoning. Despite significant progress in XAI, several challenges remain, including scalability, balancing performance with interpretability, and addressing ethical concerns, presenting opportunities for future research.

\nocite{wang2023preventing,zhang2023individual,wang2023fg2an,wang2023mitigating,chinta2023optimization,wang2024history,chu2024fairness,yin2024improving,wang2023towards,wang2024toward,chinta2024fairaied,wang2024individual1,doan2024fairness1,wang2024advancing,wang2024group,wang2024individual,yin2024accessible,wang2025fg,wang2025graph,wang2025fair,wang2025towards,yin2025digital,chinta2025ai,wang2025fdgen,wang2025Fairness,wang2025Redefining,zhang2019faht,zhang2024ai,zhang2022longitudinal,zhang2023censored,zhang2025fairness,zhang2022fairness,wang2024towards,wang2025AI,chu2024history,saxena2023missed,zhang2019fairness,zhang2020flexible,zhang2020online,zhang2020learning,zhang2021farf,zhang2021fair,zhang2023fairness,zhang2016using,zhang2018content,zhang2021autoencoder,zhang2018deterministic,tang2021interpretable,zhang2021disentangled,yazdani2024comprehensive,liu2021research,liu2023segdroid,cai2023exploring,guyet2022incremental,zhang2024fairness,wang2025FairnessT,zhang2025online,yinAMCR2025,Wang2025Unified,ijcai2025p64,ijcai2025p63,zhang2025datasets}

\bibliography{sn-bibliography}

\end{document}